# A New Segment Routing method with Swap Node Selection Strategy Based on Deep Reinforcement Learning for Software Defined Network

Miao Ye [1,4], Jihao Zheng [1,4], Qiuxiang Jiang [2], Yuan Huang [3], Ziheng Wang [1,4], Yong Wang [1,4*]

[1] School of Information and Communicatio, Guilin University of Electronic Technology, Guilin, China,

[2] School of Optoelectronic Engineering, Guilin University of Electronic Technology, Guilin, China,

[3] School of Electrionic and Automation, Guilin University of Electronic Technology, Guilin, China,

[4] Guangxi Engineering Technology Research Center of Cloud Security and Cloud Service, Guilin University of Electronic Technology, Guilin, China,

**Correspondence:** Yong Wang (ywang@guet.edu.cn)

## ABSTRACT

The existing segment routing (SR) methods need to determine the routing first and then use path segmentation approaches to select swap nodes to form a segment routing path (SRP). They require re-segmentation of the path when the routing changes. Furthermore, they do not consider the flow table issuance time, which cannot maximize the speed of issuance flow table. To address these issues, this paper establishes an optimization model that can simultaneously form routing strategies and path segmentation strategies for selecting the appropriate swap nodes to reduce flow table issuance time. It also designs an intelligent segment routing algorithm based on deep reinforcement learning (DRL-SR) to solve the proposed model. First, a traffic matrix is designed as the state space for the deep reinforcement learning agent; this matrix includes multiple QoS performance indicators, flow table issuance time overhead and SR label stack depth. Second, the action selection strategy and corresponding reward function are designed, where the agent selects the next node considering the routing; in addition, the action selection strategy whether the newly added node is selected as the swap node and the corresponding reward function are designed considering the time cost factor for the controller to issue the flow table to the swap node. Finally, a series of experiments and their results show that, compared with the existing methods, the designed segmented route optimization model and the intelligent solution algorithm (DRL-SR) can reduce the time overhead required to complete the segmented route establishment task while optimizing performance metrics such as throughput, delays and packet losses.

## 1. Introduction

In today's internet era, network communication has become an indispensable component of human society. With the rapid development of 5G and the Internet of Things, the amount of network traffic has grown rapidly in recent years. Therefore, designing an efficient routing path planning method, balancing the large amount of generated network traffic, avoiding network congestion, and improving network performance are crucial tasks (He et al. 2022).

In traditional network architectures, the data forwarding and control management functions are tightly coupled, and network device switches are responsible for packet forwarding, routing decision-making and network control; network configuration and management schemes are dispersed across different network devices. Due to the rapid development of network technology and application, problems such as the presence of highly heterogeneous types of devices underlying networks, the continuous expansion of network scales and various network protocols have been brought about by the constant emergence of new service requirements, and they have gradually complicated routing strategies(Dong et al. 2024). Thus, the traditional network architecture is difficult to flexibly adjust according to real-time network requirements or states. All of these issues bring serious challenges to the deployment, configuration and management processes involved in the traditional network architecture.

The advent of a software-defined network (SDN) has simplified the complicated traditional network configuration procedure and separated the control plane from the data plane. The controller of an SDN can obtain the global information of the network and control network devices through software programming to realize centralized network management and control, provide a unified network resource allocation scheme, and exhibit improved resource utilization and business flexibility (Kumar et al. 2023). The switch in the data plane is only responsible for forwarding network packets. The architectural design for decoupling the control plane from the data plane is conducive to the configuration and flexible deployment of network devices (Kang and Cho, 2022).

Compared with the traditional network architecture, an SDN implements relatively flexible route configuration and forwarding management processes, but the delivery of SDN routing policies requires the installation of corresponding flow table entries for all network nodes. When the network status changes and routes are adjusted, frequently updating the flow table for all network nodes increases the burden imposed on the control plane. The time delay of the streaming table update process also affects the overall performance of the network. Moreover, the SDN switch has limited ternary content-addressable memory (TCAM) capacity, storing numerous stream tables depletes resources, and the presence of numerous streaming table entries increases the latency of the switch in terms of processing each packet, thus affecting the performance of the network. The emergence of segment routing (SR) provides more flexible and efficient network management and control methods for SDN architecture. SR directly embeds path information into the packet, reducing the pressure on the SDN controller to send the flow table frequently and the network's dependence on the SDN controller and improving the forwarding efficiency of network devices (Filsfils et al. 2018). Instead of introducing stream entries for each node in the data plane, the SR process only needs to maintain all routing policies on the entry node (or segmented swap node). The entry switch directly specifies the transmission path of the packet across the network by inserting a routing list (usually called the segment list) in the header of the packet. The transit node only considers forwarding the top SID packet in the list. SR, as a way to control the packet forwarding process, effectively reduces the quantity of flow table entries contained in the data layer and improves the flexibility and scalability of the network (Abdullah et al. 2019).

Unlike the original SDN architecture, which requires a flow table to be configured for each SDN switch node, SR only needs to maintain the state of each flow on the entry node or the segmented swap node and install the segment list on these nodes. At present, there are two main ways to implement segment routing in an SDN: multiprotocol label switching (MPLS) and segment routing through IPv6 (SRv6). It should be noted that although the subsequent discussion in this paper takes MPLS as an example to study the proposed segmentation and route planning strategies, the design method is not limited to any specific implementation of SR; it also applies to the segmented routing method implemented through SRv6.

Because the label stack lengths of actual deployed MPLS devices are limited, the maximum stack depth, also known as the stack list depth (SLD), must be considered when encoding the segmented path. When the path length exceeds the SLD supported by the switch, one label stack cannot carry all the link labels, so the controller must divide the entire path into multiple label stacks (Filsfils and Michielsen, n.d.; Guedrez et al. 2017). A special label is used to "glue" adjacent label stacks together, connecting multiple label stacks in an end-to-end manner to identify a complete label switching path (LSP). This special label is called a swap label, and the node where the swap label is located is the swap node according to the literature (Ali et al. 2017). The controller assigns the swap labels to the swap nodes, attaches the swap labels to the bottom of the upstream label stack of the LSP, and associates the swap labels with adjacent downstream label stacks. Unlike link labels, swap labels cannot identify links. When a packet is forwarded to a swap node based on the upstream label stack of the LSP, the new label stack replaces the swap label according to the association between the swap label and the downstream label stack, and the model continues to forward packets that are downstream of the LSP.

The existing segmented routing methods usually first determine the shortest route path and then segment the previously determined route path with the maximum segment list depth as a constraint. According to the segmentation results, the controller selects a swap node to allocate the partition label and deliver the flow table to (Bhatia et al. 2015; Zhou et al. 2019). In this segmentation mode, the nodes in the path with low communication costs for the controller are not used as swap nodes to divide the labels. Although the controller calculates its route from the source node to the destination node with the lowest cost based on the global network status information, the cost incurred by the controller for issuing segment exchange labels to swap nodes may be very high. Therefore, the overall time cost of the flow table delivered by the controller is not reduced to the greatest extent, and the overall time cost of establishing the segmented route is increased. Therefore, when establishing SR paths that satisfy the set segment list depth constraints, the communication cost of the route path from the source node to the destination node and the communication cost of selecting a suitable swap node for considering the flow table delivered by the controller to the swap node must be addressed.

To build an efficient segmented routing network, it is necessary not only to establish an optimal path planning model considering the swap node selection strategy, but also to design a solution method that adapts to the high-speed dynamic changes exhibited by the network state. In the problem solved in this paper, the establishment of the optimal routing path from the source node to the destination node is inseparable from complex and multidimensional network state information, and the optimal swap node-based selection method is also inseparable from the network delay. Such complex network state information brings great difficulties to path planning and swap node selection. The routing algorithms implemented under the traditional network architecture, including the shortest path method (Oki et al. 2015; Tao et al. 2021), are difficult to adapt to dynamic network state changes because they cannot make full use of global network information and have slow convergence rates and long response times. Heuristic routing methods, including genetic algorithms (Bhowmik and Gayen, 2023) and the particle swarm optimization algorithm (Kabiri et al. 2022), have strong global optimal solution acquisition capabilities and only require simple iteration operations, which are easy to implement. However, due to the large number of required computations, these methods face the problem of slow convergence. In recent years, artificial intelligence technology has rapidly developed, and some intelligent solving methods have exhibited great advantages in terms of addressing highly complex optimization problems under nonlinear and complex constraints. Many studies have begun to apply artificial intelligence methods to solving route optimization problems (Casas-Velasco et al. 2022; Ye et al. 2024). Among them, deep reinforcement learning (DRL) is a data-driven artificial intelligence method that can handle high-dimensional state spaces with large numbers of features or complex representations by using deep neural networks. Deep reinforcement learning agents can independently learn strategies by interacting with the environment, which makes DRL more advantageous for handling complex and dynamically changing route optimization problems (Yao et al. 2020).

Therefore, to solve the problems that the above-mentioned existing methods increase the time cost of establishing segment routes in stages under segment list depth constraints and have a weak ability to adapt to high-speed dynamic network changes, on this paper, an intelligent adaptive SDN-based segmentation routing algorithm based on deep reinforcement learning (DRL-SR) is designed. Compared with the existing segmented routing methods, which first determine the target routing path and then select the cohesive node with the maximum segment list depth as the constraint condition for segmenting the path into multiple label stacks, this paper establishes an optimal model for both path planning and cohesive node selection tasks and designs a deep reinforcement learning routing algorithm that adapts to the high-speed dynamic network state changes. Under the segment list depth constraint, the path planning can be completed, and the most suitable nodes with low communication delays between the controller and the path can be selected as the adherent nodes so that the controller can minimize the time cost of sending the flow table to these optimal adherent nodes. First, the control plane collects global network traffic information under a software-defined network architecture and generates traffic matrix consisting of link bandwidth, link delay, and packet loss rate information. Secondly, the network topology, label stack depth and generated traffic matrix are designed as the agent environment in deep reinforcement learning. For the agent, the action selection strategy is designed by considering not only path planning for selecting the next hop node but also whether the newly added node is selected as the adherent node. Finally, the agent can continuously learn and adapt to the dynamic changing network state, generate the best forwarding path under the guidance of the reward function, and flexibly select the adhesion nodes in the path with low communication delay to the controller to divide the label stack to optimize the performance of the network and accelerate the establishment of segmented routes.

The innovations of this paper are as follows:

1) In contrast to the existing segmented routing method, which must first determine the target routing path and then determine the swap node according to the depth of the maximum segment list in dividing multiple label stacks, this paper establishes a combinatorial optimization model under the SDN architecture that can simultaneously obtain the path planning scheme and the optimal segmented route-based swap node selection strategy, considering the time cost of delivering the flow table from the controller to the optimal swap node. The speed of flow table delivery and the performance of the segmented route are maximally improved. In addition, we provide a mathematical proof that the combinatorial problem designed in the optimization model is NP-hard.

2) To solve the designed NP-hard combinatorial optimization problem, considering the optimized segmented routing model and the weak ability of current routing methods to adapt to network state changes, an intelligent solution algorithm based on deep reinforcement learning (DRL-SR) is designed. On the basis of the collection of link residual bandwidths, transmission delays, link packet loss rates and communication delays between controllers and switches under the SDN architecture, the intelligent scheme can constantly learn and adjust its update strategy in a highly dynamic network environment. The routing and forwarding paths with higher bandwidths, lower delays and lower packet loss rates are determined. Additionally, a node on a path with a shorter communication delay than the controller is selected as the swap node, and an efficient segmentation route is established.

3) The designed reinforcement learning algorithm uses the SAC algorithm in the AC framework as the core framework, and the traffic matrix consisting of global network state information is combined with the network topology and the SR label stack depth to form the state space of the agent in deep reinforcement learning. The agent not only designs an action selection strategy to choose the next node in path planning but also designs an action selection strategy for determining whether the newly added node is a swap node. This action selection strategy is based on the different actions taken by the agent in different state spaces. A reward function, which considers factors such as the optimization of the forwarding path and the time cost of the controller sending the flow table to the swap node, is designed.

4) The results of a series of experiments conducted for multiple real network topologies show that, compared with the existing segmented routing method that selects swap nodes to divide label stacks according to the maximum segment list depth, the designed DRL-SR method can optimize the throughput, delay and packet loss rates and reduce the delivery time required for the flow table in the SDN to establish routes more quickly.

The rest of this article is organized as follows. The related work is described in section 2. Section 3 analyzes the addressed problem and introduces the SDN-based intelligent segmented routing scheme. The DRL-SR algorithm is introduced in detail in Section 4.

Section 5 describes the experimental setup and performance evaluation results. Section 6 introduces the conclusions and future work related to this paper.

## 2. Related work

In this section, we discuss the route optimization method and the related work of SR. The advantages and disadvantages of different route calculation methods in network optimization are analyzed, and the existing work related to segmented routing in network optimization is described.

Routing optimization method: Determining the optimal routing path in a real-time dynamically changing network is highly important for optimizing network performance. Currently, there are multiple classical routing path optimization methods. Derbel et al. (Derbel et al. 2012) highlighted a genetic algorithm (GA) combined with iterative local search (ILS), which strengthens the search space and addresses the concern that the solutions generated by the GA are prone to falling into local optimality. Zhang et al. (Zhang et al. 2018) proposed a combined GA– and bacterial foraging optimization algorithm to select the optimal path; this algorithm can more easily determine the extreme value and optimal path and compensate for the poor accuracy and local optimization of the GA. Parsaei et al. (Parsaei et al. 2017) modeled a quality of service (QoS) protocol as a constrained shortest path (CSP) linear programming problem and proposed a solution method based on the ant colony algorithm. Truong Dinh et al. (Truong Dinh et al. 2020) proposed a heuristic traffic engineering method based on multi-path forwarding and inter-path traffic exchange, which determines the initial path with the lowest cost selection from k available paths and then triggers heuristic redynamic selection of the optimal path according to the path load and flow properties.These methods also have the limitations of heavy computations and poor adaptability to high-speed changes in the network state.

With the continuous development of computer science, intelligent optimization algorithms can better handle and adapt to complex and high-dimensional dynamic network environments, so intelligent algorithms have also achieved good development in routing path optimization. Yanjun et al. (Yanjun et al. 2014) proposed a meta-layer framework based on supervised machine learning to solve dynamic routing problems in real time. Multiple machine learning modules are constructed in the meta-layer, for which the training set consists of the input of the heuristic algorithm and its corresponding output. After the training process, the meta-layer directly and independently yields similar heuristic results, thereby replacing the time-consuming heuristic algorithm and effectively improving the network performance. Mao et al. (Mao et al. 2021), aiming to address the lack of adaptive capability of routing policies with maximum or minimum metrics in a software-defined communication system (SDCS), designed a convolutional neural network (CNN) to intelligently calculate the path on the basis of the input real-time traffic trajectory to improve the adaptability of the CNN to changing traffic patterns. To achieve proper input and output characterization of heterogeneous network traffic, Kato et al. (Kato et al. 2017) proposed a supervised deep neural network system approach to improve the performance of heterogeneous network traffic control. These machine learning methods effectively improve network performance, but machine learning requires a substantial amount of labeled data for training, which is difficult to obtain in complex dynamic networks. The accuracy of datasets also affects the accuracy of the system.

Compared with machine learning, which usually requires static datasets for training and has difficulty adapting to real-time changes, RL can learn, optimize and adapt to a dynamic environment; thus, many excellent reinforcement learning routing optimization methods have emerged. Duong et al. (Duong and Binh, 2022) proposed an intelligent routing algorithm based on machine learning. A combination of supervised learning (SL) and RL, the algorithm predicted the performance indicators of links, including EED quality of transmission (QoT) and packet blocking probability (PBP), and Q-learning reinforcement learning was used to determine the routing target. Chen et al. (Chen et al. 2020) proposed an RL method to solve the traffic engineering problem of throughput and delay in SDNs. Huang et al. (Huang et al. 2022) used a GRU model to predict the traffic information of an SDN and used the K path from source to destination calculated by the Dijkstra algorithm as the action of agent selection to dynamically search for the optimal routing strategy. Liu et al. (Liu et al. 2021) proposed a routing scheme with a resource reorganization state, which uses a deep Q-network (DQN) and a deep deterministic policy gradient (DDPG) to construct DRL-R that optimizes the allocation of network resources for traffic, constantly interacts with the network, and performs adaptive routing according to the network state. Compared with the traditional routing algorithm and machine learning routing optimization algorithm, the RL method greatly improves network performance and can dynamically adjust the routing strategy according to the network state in the dynamic network environment, showing significant advantages.

Related work for SR: SR uses the characteristics of SDN architecture to separate the control plane and data plane and directly inserts path information from the source node into the packet header via source routing, which improves the efficiency and flexibility of packet forwarding. To date, many studies on SR have been conducted. SR has been confirmed to reduce the minimizing forwarding table size (FTS) of a switch. Anbiah and Sivalingam (Anbiah and Sivalingam, 2021) studied the minimizing FTS problem under a given flow set and SLD limitation. Two different heuristic solutions were proposed. Li et al. (Li et al. 2016) used source routing to replace the table lookup-based approach in traditional SDNs, which improved the efficiency of the forwarding plane and significantly reduced the path establishment traffic delay. Li and Hu (Li and Hu, 2020) proposed an efficient flow routing scheme based on SR, which aggregates numerous flows into a small number of flow items according to the degree of overlap of flow paths to achieve path aggregation and solve the problem of a shortage of flow table resources in SDN switches. Dong et al. (Dong et al. 2017) proposed an efficient forwarding scheme based on MPLS source routing to effectively control the tradeoff between traffic overhead and bandwidth overhead. Tulumello et al. (Tulumello et al. 2023) proposed a Micro SID solution for efficiently representing Segment identifiers in SRv6, minimizing the impact on the MTU (maximum transport unit) when carrying a large number of segments in an IPv6 header.

Cianfrani et al. (Cianfrani et al. 2017) proposed an SR domain (SRD) architecture solution to ensure correct interworking between IP routers and SR nodes and optimize the maximum link utilization when only some nodes have SR capability. Guo et al. (Guo et al. 2021) optimized the shunting ratio of SR nodes in a centralized online manner to improve network performance under the dynamic traffic requirements of hybrid SR networks.

Zhang et al. (Zhang et al. 2022) introduced segmented routing for the first time in Wireless Mesh Networks (WMNs), and proposed an online primitive dual algorithm to ensure the performance lower bound in the worst case. Aureli et al. (Aureli et al. 2022) adopted the source routing function of SR in the framework based on deep reinforcement learning. The agent selects reroute operations according to the link load to move traffic from the overloaded link to the alternate path, which can achieve link traffic balancing without affecting the global maximum link utilization.

SR relies on label stacking and does not require signaling protocols. This method greatly simplifies the network operation of the transport node but introduces scalability issues with entry nodes and packet overhead. Owing to the constraint of SLD, labels cannot be inserted into packets indefinitely. When the route length exceeds the SLD, optimization of the label stack is a crucial problem in SR. A specific algorithm is needed to efficiently compute the label stack for a given path. Giorgetti et al. (Giorgetti et al. 2015) proposed two SR label stack computing algorithms that guarantee the minimization of the label stack depth. Dugeon et al. (Dugeon et al. 2017) combined the capabilities of an SDN controller and a path coding engine to reduce the size of the label stack to represent SR paths. Guedrez et al.(Guedrez et al. 2016) used the existing IGP shortest path in the network to represent the minimum label stack of SR-MPLS paths according to MSD constraints, reducing the impact of MSD and ensuring the path diversity of SR in the network. Lazzeri et al. (Lazzeri et al. 2015) proposed an efficient segment list coding algorithm to ensure optimal path calculation and minimize the SLD in SR networks. Utilizing the network programmability provided by OpenFlow, Huang et al. (Huang et al. 2018) proposed an improved SR structure for the data plane, which reduced the overhead of extra stream entry and label space, and designed a new path coding scheme to minimize SLD under given maximum constraints, accounting for multiple types of overhead. Moreno et al. (Moreno et al. 2017) proposed heuristic methods to perform segment list calculations accurately, using a very limited number of stacked tags to achieve a very efficient TE scheme.

Currently, most studies on SR focus on implementation methods of SR path coding via SLD and effective TE solutions in SR networks. However, in these works, the optimization methods are only applicable to previously determined paths and rarely consider methods of selecting swap nodes when optimizing SR to reduce the flow table delivery time of SDN controllers. To reduce the time cost of flow table delivery when adjusting route switching in the SDN and accelerate the establishment of segmented routes, this paper designs a segmented route based on a DRL algorithm to overcome the fact that segmented route coding depends on previously determined paths and to realize a segmented route that can quickly establish the optimal routing path in a dynamic changing network.

## 3. Optimized model of the segmented routing path planning and swap node selection strategy

In this section, we introduce an optimization model of the SDN architecture that can address both the path planning strategy and segmented route swap node selection.

The SDN controller calculates the forwarding path from the source node to the destination node. The calculated path integrates the link labels of the entire path according to the link labels of the topology to generate a label stack. When the label stack depth exceeds that supported by the forwarder, one label stack cannot carry all the link labels. Therefore, the controller has to divide the entire path into multiple label stacks. Finally, the controller passes the label stack to the entry node and the swap node. The transponder establishes a segmented route on the basis of the label stack issued by the controller. Figure 1 illustrates the label delivery process executed after the routing path is obtained and the procedure through which the switch forwards data based on the label information.

Control layer label issuance process: As shown in Figure 1, the controller calculates the path from node $A$ to node $J$, i.e., $A \rightarrow B \rightarrow C \rightarrow D \rightarrow F \rightarrow H \rightarrow I \rightarrow J$, and the path length from node $A$ to node $J$ is 7. Assuming that the depth of the current label stack is 4, the path length is greater than the depth of the label stack. The labels need to be divided into three label stacks $\{1001,1003,1004,100\}$, $\{1009,1012,1013,101\}$, and $\{1015\}$, where 100 and 101 are swap labels that are associated with $\{1000,1012,1013,101\}$ and $\{1015\}$, respectively. The other labels are link labels.

The controller sends label stack $\{1001,1003,1004,100\}$ to entry node $A$, swap label

100 and label stack {1009,1012,1013,101} to swap node $D$, and swap label 101 and label stack {1015} to swap node $I$.

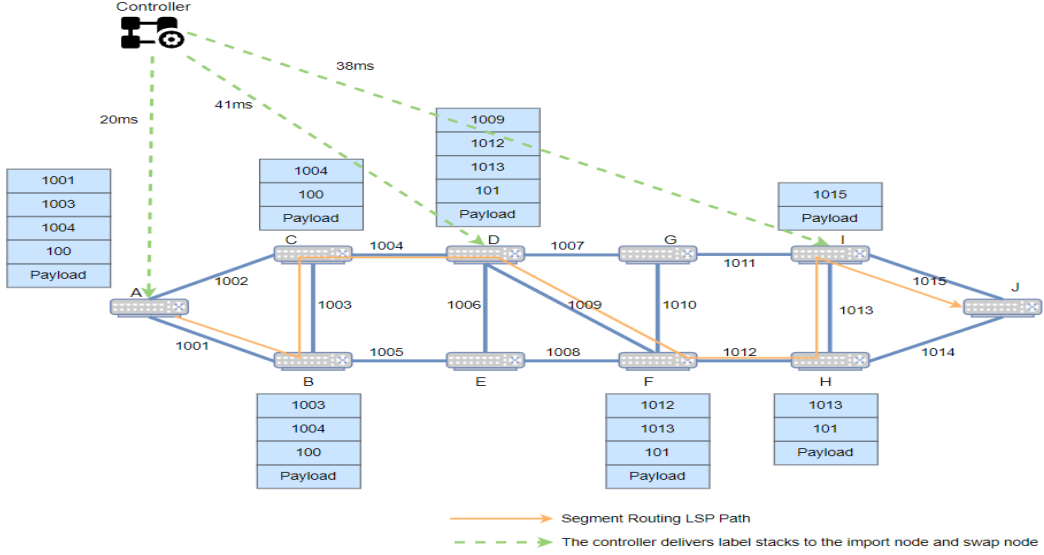

**Figure 1**. Process of establishing a segmented route.

The data layer forwarding process is as follows:
1) The entry node $A$ adds the label stack {1001,1003,1004,100} to the data packet and matches the link with label 1001 on the top of the stack to find the corresponding forwarding interface $A \rightarrow B$ link and ejects the label 1001 1001 . The packet carries the label stack {1003,1004,100} and forwards it to the downstream node $B$ through link $A \rightarrow B$.
2) After receiving the packet, node $B$ forward the data packet with tag stack {1004,100} to node $C$ in the same way.
3) After receiving the packet, node $C$ matches the link according to label 1004 on the top of the stack, finds the corresponding outbound interface as the $C \rightarrow D$ link, and ejects the label 1004. The packet carries the tag stack {100} and is forwarded to the downstream node $D$ through the $C \rightarrow D$ link.
4) After receiving the packet, the swap node $D$ identifies label 100 at the top of the stack as the swap label, switches the swap label 100 to its associated label stack {1009,1012,1013,101}, matches the new label 1009 at the top of the stack, finds the corresponding outbound interface as the $D \rightarrow F$ link, and ejects the label 1009 . The packet carries the tag stack {1012,1013,101} and is forwarded to node $F$ via the $D \rightarrow F$ link.
5) Nodes $F$、$H$、$I$, as above, forward data packets to egress node $J$ according to the label stack.
6) The packet received by egress node $J$ does not carry labels and is forwarded by searching the routing table.

Assume that along path $A \rightarrow B \rightarrow C \rightarrow D \rightarrow F \rightarrow H \rightarrow I \rightarrow J$, the latency values between the nodes and controllers are $20, 25, 28, 41, 35, 27, 38, 27$. In Figure 1, nodes $D$ and $I$ are selected as swap nodes. The establishment time of the segmented route is $\max(20,41,38) = 41$. Without changing the path, when the swap nodes are changed and nodes $C$ and $H$ are selected as swap nodes (as shown in Figure 2).

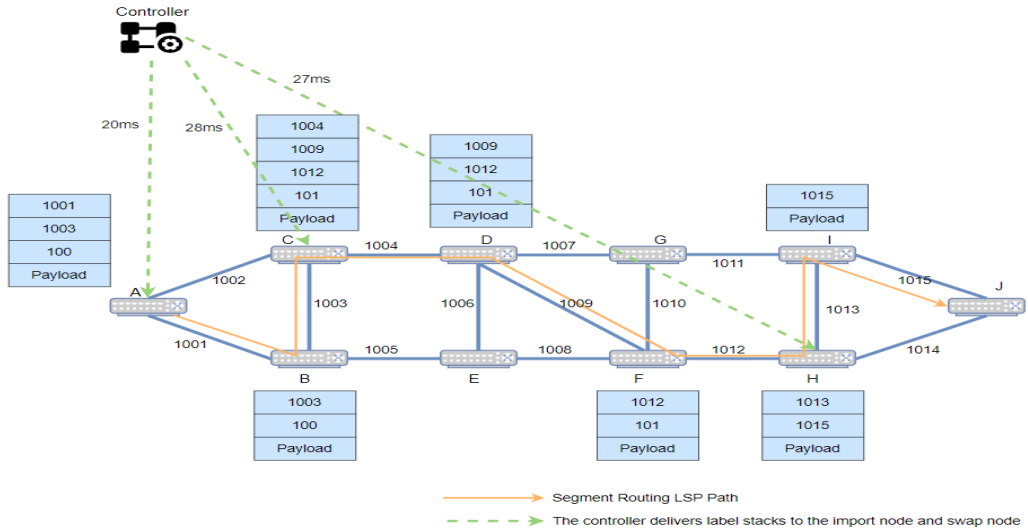

**Figure 2.** Selection of the swap node.

The establishment time required for the segmented route is $\max(20,28,27) = 28$. The speed for this route is faster than that attained when selecting nodes $D$ and $I$ as the swap nodes.

For the segmented routing path optimization problem proposed above, the swap node selection strategy is established as the following optimization model.

Suppose that the network topology is abstracted as an undirected, weighted connected graph $G = (V, E, W)$, where $V$ is the set of nodes, $E$ is the set of edges, and $W$ is the weight factor for the edges. For the edge $e_{ij} \in E$ between node $i$ and node $j$ in the graph, there is generally a weight $w(e_{ij})$. In this paper, $w(e_{ij})$ is considered a mapping value for three performance indicators, namely, the bandwidth $bw_{ij}$, delay $delay_{ij}$, and packet loss rate $loss_{ij}$:

$$w(e_{ij}) = w(bw_{ij}, delay_{ij}, loss_{ij}) \quad (1)$$

Assume that the path from the source node $src \in V$ to the destination node $dst \in V$ needs to be divided into segmented path of $l$ segments:

$$P = <p_1, p_2, \ldots, p_l> \quad (2)$$

where $P$ and $l$ are both variable values that we need to determine, $P$ involves determining how to reach the planned path from the source node to the destination node, $l$ involves determining how many segments the path $P$ is divided into. $p_i$ is the segmented path component of section $i$ in path $P$. When $i = 1$ is the first segmented path from the source node, $p_i$ is obviously a subpath of $P$, which can be expressed as $p_i = (V_i, E_i, W_i)$. $V_i$ represents all nodes along the segmented path $p_i$. $E_i$ represents all edges of the segmented path $p_i$. $W_i$ represents the weights of all sides in the segmented path $p_i$.

$$V_i = \{v_{i,1}, v_{i,2}, \ldots, v_{i,m_i}\}, i = 1, 2, \ldots, l \quad (3)$$

$$E_i = \{e_{v_{i,j}v_{i,j+1}}\}, v_{i,j}, v_{i,j+1} \in V_i \quad (4)$$

$$W_i = \{w(e_{v_{i,j}v_{i,j+1}})\}, v_{i,j}, v_{i,j+1} \in V_i \quad (5)$$

$m_i = |v_i|$ indicates the number of nodes contained in the segmented path $p_i$ for segment $i$, which satisfies the condition that the depth of the label stack must be less than $M$, that is, $m_i \leq M$. $M$ is a constant that represents the label stack depth, which is determined by the hardware configuration conditions in advance. $v_{i,j}$ is the JTH node along the $i$ th segmented path $p_i$; obviously, $v_{1,1}$ is the entry source node of the segmented route. $v_{i,1}(i \neq 1)$ is the first node on the $i$ th segmented path $p_i$ belonging to the swap node. $e_{v_{i,j}v_{i,j+1}}$ is the edge between the $J$ th node of the segmented path $p_i$ and the $j + 1$ th node of $p_i$.

Both $P$ and $l$ are variable values that we want to determine. When $P$ and $l$ take corresponding values of $P^*$ and $l^*$, respectively, the cost $f(P^*)$ corresponding to a certain aspect of the piecewise path $P$ is minimized. The cost function $f(P)$ is designed as follows.

The remaining bandwidth of $P$, $bw$: The minimum remaining bandwidth from the source node $src$ to the destination node $dst$ can be expressed as the minimum bandwidth across all links, so it can be defined as:

$$bw = \min_{e_{ij} \in E_i}(bw_{ij}) \quad (6)$$

where $bw_{ij}$ is the remaining bandwidth of the link between node $i$ and node $j$.

Total delay of path $P$, $delay$: This value represents the sum of delays of all links in $P$, as defined in formula (7):

$$delay = \sum_{e_{ij} \in E_i} delay_{ij} \quad (7)$$

where $delay_{ij}$ is the delay of the link between the node $i$ and node $j$.

Packet loss rate of path $P$: This is the product of packet loss rates for all links on path $P$:, as shown in formula (8):

$$loss = 1 - \prod_{e_{ij} \in E_i}(1 - loss_{ij}) \quad (8)$$

where $loss_{ij}$ is the packet loss rate of the link between node $i$ and node $j$.

$cdelay$, the delay of all the swap nodes in the segmented path $P$ for completing the delivery of the flow table: This value indicates the delay required for the controller to complete the delivery of the label stack from the source nodes $v_{1,1}$ and all the swap nodes $v_{i,1}$. When establishing a segmented route, the controller needs to deliver label stacks for the entry node and all the swap nodes. The entire process of delivering label stacks is complete only when the task of delivering each entry node (or the swap node) in all segments is completed. Therefore, the delay cost incurred when delivering a flow table is the maximum time cost of $l$ swap nodes, which can be expressed as:

$$cdelay = \max(cdelay_{v_{1,1}}, cdelay_{v_{2,1}}, \cdots, cdelay_{v_{l,1}}) \quad (9)$$

$cdelay_{v_{i,1}}$ indicates the delay required for the controller to deliver the label stack to $v_{i,1}$.

When packets are transmitted from the source node to the destination node along path $P$, the maximum $bw$, minimum $delay$, and $loss$ are required. At the same time, to complete the flow table delivery as soon as possible, the label stack depth constraint $m_i \leq M, i = 1, \ldots, l$. Under this condition, the delay cost formula (9) needs to be minimized:

$$\min bw = \min_{e_{ij}\epsilon E_i}(bw_{ij})$$

$$\min delay = \sum_{e_{ij}\epsilon E_i} delay_{ij}$$

$$\min loss = 1 - \prod_{e_{ij}\epsilon E_i}(1-loss_{ij}) \quad (10)$$

$$\min cdelay = \max(cdelay_{v_{1,1}}, cdelay_{v_{2,1}}, \cdots, cdelay_{v_{l,1}})$$

$$m_i \leq M, i = 1, \ldots, l$$

The optimization model established above is actually a multiobjective optimization problem: $[\max bw, \min delay, \min loss, \min cdelay]$. Multiobjective optimization means that when multiple objectives are to be achieved in a certain scenario, due to the inherent conflicts between the objectives, the optimization of one objective usually comes at the cost of the deterioration of other objectives. Usually, the only optimal solution is not obtained; instead, a set containing many optimal solutions is obtained on the multiobjective Pareto front. In engineering applications, the required solutions are selected to optimize the allocation of resources. Another common approach is to carry out linear weighting to convert multiple-objective optimization problems into single-objective optimization problems. In this paper, four indices, the bw, delay, loss and cdelay, are normalized to [0,1]. Then, the cost function $f(P)$ designed for the segmented path $P$ under the condition that the above multiple objectives are equally important is as follows:

$$\min f(P) = \beta_1(bw-1) + \beta_2(1-delay)$$
$$+ \beta_3(1-loss) + \beta_4(1-cdelay)$$

$$\text{s.t.} \ P = <p_1, p_2, \ldots, p_l>$$

$$P_i = (V_i, E_i, W_i)$$

$$V_i = \{v_{i,1}, v_{i,2}, \ldots, v_{i,m_i}\}, i = 1, 2 \ldots, l \quad (11)$$

$$E_i = \{e_{v_{i,j}v_{i,j+1}}\}, v_{i,j}, v_{i,j+1} \in V_i$$

$$W_i = \{w(e_{v_{i,j}v_{i,j+1}})\}, v_{i,j}, v_{i,j+1} \in V_i$$

$$m_i \leq M, i = 1, \ldots, l$$

The variables that are subject to optimization in the established optimization model (11) are $P$ and $l$. These are discrete variables, with $P$ defined as follows: $P = <p_1, p_2, \cdots, p_l>$. It is therefore evident that (11) belongs to the combinatorial optimisation problem, with $P$ representing a path between the starting point and the end point of the given path, $l$ representing the division of this path into part $l$, and the value range of l being an integer between 1 and $n$. Furthermore, the length $m_i$ of each subpath after $l$-partition of the path $P$ does not exceed the given stack depth constant $M$, $M$, $i = 1, \ldots, l$. The subsequent section provides a comprehensive proof that the optimization model (11) is an NP-hard combinatorial optimization problem.

**THEOREM 1**: Optimization model (11) is an NP-hard combinatorial optimization problem.

**Proof:** If there is no constraint $m_i \leq M$, a path of length $n$ can be subdivided into at most $n$ segments and at least 1 segment; when it is divided into $l$ segments, the number of ways to divide it is $C_{n-1}^{l-1}$, where $l = 1, \ldots, n$. If we denote $f(n)$ as the number of ways in which a path of length $n$ can be divided into segments, with $l$ representing the various lengths of these segments ($l = 1, \ldots, n$), then we have:

$$f(n) = C_{n-1}^0 + C_{n-1}^1 + \ldots + C_{n-1}^{n-1} = 2^{n-1} \quad (12)$$

When the constraint $m_i \leq M$ holds, that is, the length of each subpath $m_i$ must be no greater than the given stack depth constant $M$, the following two scenarios can be considered:

Case 1: The path length $n$ satisfies $1 \leq n \leq M$, which is equivalent to the case in which there is no constraint $m_i \leq M$; we have:

$$f(n) = 2^{n-1}, 1 \leq n \leq M. \quad (13)$$

Case 2: The path length $n$ satisfies $n > M$, the length of the first divided segment satisfies $m_1 = 1, \cdots, M$, and the remaining segment lengths may be any value in the range $n-1, \cdots, n-M$; we have:

$$f(n) = f(n-1) + f(n-2) + \cdots + f(n-M) \quad (14)$$

To demonstrate that optimization model (11) is an NP-hard combinatorial optimization problem, the solution method of the Fibonacci sequence generating function is employed to construct a generating function $F(z)$ for $(n)$:

$$F(z) = f(0) + f(1)z + f(2)z^2 + \cdots = \sum_{m \geq 0}^{\infty} f(m)z^m \quad (15)$$

$$\begin{aligned} zF(z) &= f(0)z + f(1)z^2 + \cdots + f(M-1)z^M + f(M)z^{M+1} + \cdots \\ z^2 F(z) &= f(0)z^2 + \cdots + f(M-2)z^M + f(M-1)z^{M+1} + \cdots \\ &\vdots \\ &\vdots \\ z^M F(z) &= f(0)z^M + f(1)z^{M+1} + \cdots \end{aligned} \quad (16)$$

$$(1-z-\cdots-z^M)F(z) = f(0) + [f(0)-f(1)]z + \cdots + [f(M)-f(M-1)-\cdots-f(0)]z^M + \cdots \quad (17)$$

To derive Equation (16), it is necessary to multiply $z, z^2, \cdots z^M$ separately on the two sides of Equation (15). Equation (17) is obtained by subtracting both sides of Equation (15) from both sides of Equation (16). To do this, we need to extend $f(n)$ to the case of $n = 0$. From Equation (14), the following can be obtained:

$$f(0) + f(1) + \cdots + f(M-1) = f(M) \quad (18)$$

With Equation (13), we can obtain $f(1) = 1, f(2) = 2, f(3) = 4, \cdots, f(M) = 2^{M-1}$. Substituting these values into Equation (18), we have $f(0) = 1$. On the basis of $f(0) = 1$, when $n \leq M$, $f(n)$ satisfies:

$$f(n) = f(0) + f(1) + \cdots + f(n-1). \quad (19)$$

According to Equations (14) and (19), Equation (17) can be simplified as:

$$(1 - z - \cdots - z^M)F(z) = 1 \quad (20)$$

Therefore,

$$F(z) = \frac{1}{(1 - z - \cdots - z^M)} \quad (21)$$

When $M = 1$, we have:

$$F(z) = \frac{1}{(1-z)} = \sum_{m=0}^{\infty} z^m \quad (22)$$

When $M = 2$, we have:

$$F(z) = \frac{1}{(1 - z - z^2)} = \sum_{m=0}^{\infty}(z + z^2)^m = \sum_{m=0}^{\infty}\sum_{i=0}^{m} C_m^i z^{2m-i} \quad (23)$$

It has been demonstrated that f(n) is associated with the combinatorial number $C_n^i$. However, as n increases, f(n) exhibits combinatorial explosion, so finding $f(n)$ is NP-hard.

When $M > 2$, we have:

$$F(z) = \frac{1}{[1 - (z + z^2 + \cdots + z^M)]}$$
$$= \sum_{m=0}^{\infty}(z + z^2 + \cdots + z^M)^m \quad (24)$$

In this case, $f(n)$ can be reduced to the case in which $M = 2$, so it can be concluded that it is NP-hard for optimization model (11) to solve $f(n)$ for any $M > 2$ (Kleinberg and Tardos, 2005). ∎

## 4. Intelligent segmentation routing algorithm based on deep reinforcement learning

### 4.1. DRL-SR intelligent segmented routing framework considering the swap node selection strategy

The SDN framework senses network status information to obtain bandwidth, delay, packet loss rate, delay between A-nodes and controllers, etc. In DRL, the agent uses this information to learn how to build segmented routes from source nodes to target nodes, and the controller delivers flow tables to the entry nodes and swap nodes through the southbound interface. The routing policy is intelligently adjusted according to the dynamic network link information. The structure of the SDN intelligent segmentation routing strategy designed in this paper is shown in Figure 3.

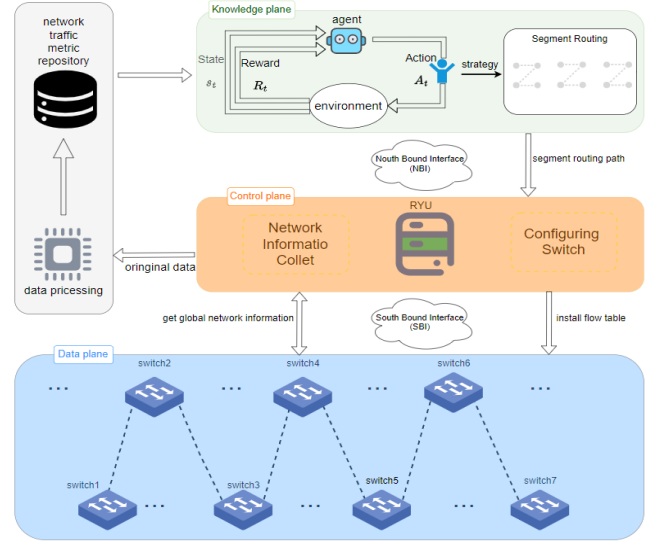

**Figure 3**. SDN intelligent segmented routing structure.

(1) Data plane: The data layer consists of the underlying switch devices, which are responsible for the actual packet forwarding operation, but their control logic is managed by the SDN controller in the control layer. The SDN controller communicates with the data layer devices via the Southbound Interface protocol (OpenFlow protocol), providing instructions on processing the data flow. The data layer provides the control layer with the original data of the switch port, including the number of packets sent by each port of the switch, $tx_p$; the number of packets received, $rx_p$; the number of bytes sent, $tx_b$; the number of bytes received, $rx_b$; the number of dropped packets sent, $tx_{drop}$; the number of dropped packets received, $rx_{drop}$; the number of wrong packets sent, $tx_{err}$; the number of wrong packets received, $rx_{err}$; the number of ports; and the duration of the bytes sent, $t_{dur}$.

(2) Control plane: The control plane is responsible for centrally controlling and managing the behavior of the entire network. The control layer consists of an RYU controller that communicates with the underlying network devices through a southbound interface and with upper-layer applications through a northbound interface. The controller periodically obtains the original data of the switch ports in the data layer and calculates the network status information discussed in this paper, including the link residual bandwidth $bw_{ij}$, link delay $delay_{ij}$, link packet loss ratio $loss_{ij}$, and flow meter installation delay $cdelay_i$, and constructs the global network view. The knowledge layer determines the optimal segmentation route according to the global network view constructed by the control layer, and the controller dynamically configures and adjusts the forwarding strategy of the network device according to the decision.

The residual bandwidth $bw_{ij}$ is the difference between the maximum bandwidth $bw_{max}$ and the used bandwidth $used_{bw_{ij}}$ of the link. The instantaneous throughput (used bandwidth $used_{bw_{ij}}$) can be calculated using $tx_b$, $rx_b$ and $t_{dur}$. The formula for residual bandwidth $bw_{ij}$ is shown in Equation (26):

$$used_{bw_{ij}} = \frac{|(tx_{bi} + rx_{bi}) - (tx_{bj} + rx_{bj})|}{t_{durj} - t_{duri}} \quad (25)$$

$$bw_{ij} = bw_{max} - used_{bw_{ij}} \quad (26)$$

$tx_{bi}$ and $tx_{bj}$ indicate the numbers of bytes received by nodes $i$ and $j$ respectively. $t_{duri}$ and $t_{durj}$ indicate the durations of the bytes sent by the ports of nodes $i$ and $j$ respectively.

The packet loss ratio $loss_{ij}$ is calculated from the number of packets sent $tx_p$ and the number of packets received $rx_p$. The formula is shown in Equation (27):

$$loss_{ij} = \frac{tx_{pi} - rx_{pj}}{tx_{pi}} \quad (27)$$

Flow table installation $cdelay_i$ uses the controller to send an echo request message with a timestamp to the switch; then, the controller parses the echo-reply message returned by the switch and subtracts the sending time of packet parsing from the current time. The round trip delays $T_{echo_{sendi}}$ and $T_{echo_{recivei}}$ between the controller and the switch are obtained, and $cdelay_i$ is the average of $T_{echo_{sendi}}$ and $T_{echo_{recivei}}$. The calculation formula is shown in Equation (28):

$$cdelay_i = \frac{T_{echo_{sendi}} + T_{echo_{recivei}}}{2} \quad (28)$$

The transmission delays $T_{lldp_{apij}}$ and $T_{lldp_{apji}}$ from the controller to the source switch, from the source switch to the destination switch, and from the destination switch to the controller can be calculated using the LLDP packet receiving time minus the packet sending time (Li et al. 2018). The link delay $delay_{ij}$ is calculated as shown in Equation (29):

$$delay_{ij} = \frac{\left(T_{lldp_{apij}} + T_{lldp_{apji}} - T_{echo_{sendi}} - T_{echo_{recivei}}\right)}{2} \quad (29)$$

(3) Knowledge plane: The knowledge layer, which makes segmented routing decisions for the controller, is the core of the intelligent SR method proposed in this paper. The knowledge layer receives the link residual bandwidth, link delay, link packet loss rate, and flow table installation delay sent from the control layer and performs min–max normalization processing on these parameters to form a traffic matrix. These traffic matrices are used to train the DRL agent. Once the agent obtains a convergent reward value after training, the traffic matrix of each moment is used as the input of DRL. The agent outputs the optimal label stack from the source node to the destination node in the current state.

The controller delivers the label stack to the corresponding entry node and the swap node according to the output of the agent, completing the establishment of the segmentation route. The process of the agent building the segmented routing label stack is described in detail in Section 4.2 Algorithm Design.

(4) Application plane: The application layer contains various network applications and services, which interact with the network through the API interface provided by the SDN controller. Common SDN applications include traffic engineering, security management, load balancing, and virtual networks.

*4.2. Design of the DRL-SR algorithm*

In this work, the SAC algorithm is used as the core framework, and according to the formulaic description of the SR problem in Section III, the state space, action space and reward function of the agent are designed using the global network topology and link state information.

(1) State space design: As shown in Figure 4, to facilitate the input of the neural network, the calculated link state information, current node position and label stack depth obtained from the data layer by the control layer through the southbound interface are converted into a six-channel matrix $s = [bw, delay, loss, cdelay, stack, location]$ that can be input into DRL. All the possibilities of $s$ constitute the state space $S$.

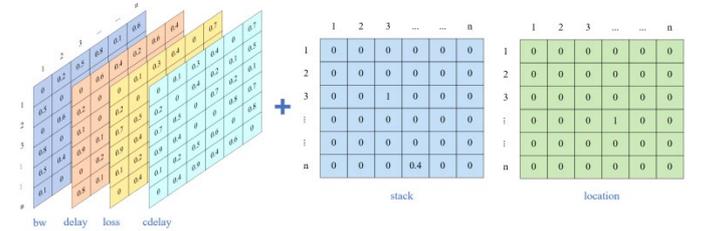

**Figure 4**. State matrix diagram of the agent.

where $bw = [bw_{ij}] \in R^{n \times n}$, $delay = [delay_{ij}] \in R^{n \times n}$, $loss = [loss_{ij}] \in R^{n \times n}$, the three matrices are the $n \times n$ adjacency matrices that we convert the data obtained from the control layer, and $n$ is the number of nodes in the topology. For example, the value $bw_{ij}$ in the $bw$ adjacency matrix represents the remaining bandwidth between node $i$ and node $j$ in the topology. If node $i$ and node $j$ are not connected, then $bw_{ij} = 0$. $cdelay = [cdelay_{ij}] \in R^{n \times n}$, the value of the diagonal in the $cdelay$ matrix, represents the delay from the corresponding node to the controller; for example, $cdelay_i$ represents the delay from node $i$ to the controller.

$stack = [stack_{ij}] \in R^{n \times n}$, where the $stack$ matrix represents the number of labels in the last stack in the current path. If the number of the last stack in the current path is $sn$, $stack_{ij} = \begin{cases} 1, i = j = sn \\ 0, \quad else \end{cases}$. $location = [location_{ij}] \in R^{n \times n}$, where the $location$ matrix represents the last node in the current path. If the last node in the current path is $vc$, $location_{ij} =$

$\begin{cases} 1, i = j = vc \\ 0, \quad else \end{cases}$. When node $vc$ is the destination node $dst$, the algorithm terminates.

(2) Action space: To enable the agent to quickly find the optimal path in the environment, determine the swap nodes in the path, and reduce the dimensionality of the actions in the action space, this paper designs an action space $A = \{a_i\}, |A| = 2n$, where $n$ is the maximum degree of all nodes in graph $G$. $a_i \in \{0,1\}^{2n}$, with the constraint $sum(a_i) = 1$, where $sum(\cdot)$ is a summation operation. That is, $a_i$ is a unique thermal coding vector, which represents the node that joins the path in the current state $s$ and determines whether the node is a swap node.

The method for determining which node joins the path via $a_i$ is as follows:

For ease of description, define $cur_{location}(s_t)$ to represent the last node in the current status path. $nei(v)$ indicates the set of neighbor nodes of node $v$. $nei(v) = \{v_0, v_1, \cdots, v_{nx-1}\}$, where $nx$ is the number of neighbor nodes of node $v$. The neighbor node set of the last node in the path of status $s$ is $nei(cur_{location}(s_t))$, and the number of neighbor nodes is $|nei(cur_{location}(s_t))|$.

$argmax(a_i)$ represents the position number of 1 in the unique thermal code $a_i$ and the remaining class of its module $n$ represents the position mapping of the neighbor node; in other words, $mod(argmax(a_i), n)$ represents the selected neighbor node number. If $mod(argmax(a_i), n) \geq |nei(cur_{location}(s_t))|$ is an invalid action. $mod(argmax(a_i), n) < |nei(cur_{location}(s_t))|$ is a valid action. If the action is valid, the neighbor node numbered $mod(argmax(a_i), n)$ is added to the path. After the effective action is executed, the current state $s_t$ is transferred to the next state $s_{t+1}$, and the $location$ matrix in the state is transferred.

$$location_{i\,j} = \begin{cases} 1, i=j=ner(cur_{location}(s_t))_{mod(argmax(a_i),n)} \\ 0, else \end{cases} \quad (30)$$

It also determines whether the node selected to join the path is a swap node. If $argmax(a_i) > n$, then the node selected to join the path is a swap node; otherwise, the node is not a swap node. The agent builds a segmented route, as shown in the following figure. If yes, the status $s_t$ changes to the next state $s_{t+1}$, and the status of the stack matrix changes. $stack$ matrix transitions are expressed as Equation (31).

$$stack_{i\,j} = \begin{cases} 1, i=j=1 \\ 0, else \end{cases} \quad (31)$$

When the nodes are not swapped, the matrix remains unchanged. Figure 5 below shows the process by which the agent selects actions to build a segmented route under different transition states.

Figure 5(a) shows the initial state, $a_t = 1$, and the set of neighbor nodes of the starting node 1 is 2,5. The first neighbor node 5 of node 1 is selected to add to path $P, P = \{5\}$. In Figure 5(b), action $a_t = 2$ selects the second neighbor node 17 of node 5 to add to path $P, P = \{5,17\}$. In Figure 5(c), with an action $a_t = 9$, the maximum degree in the graph is 6; thus, the third neighbor (node 16) of node 17 is selected to join path $P$ and is determined to be the swap node $P = \{5,17,16,100\}$. In Figure 5(d), $a_t = 1$ selects the first neighbor node 14 of node 16 to join path $P, P = \{5,17,16,100\}, \{14\}$. In Figure 5(e), action $a_t = 7$ selects the first neighbor node 8 of node 14 to add to path , and node 8 is the swap node $P = \{\{5,17,16,100\}, \{14,8,101\}\}$. In Figure 5(f), action $a_t = 4$ selects the fourth neighbor node 15 of node 8 to add to path $P, P = \{5,17,16,100\}, \{14,8,101\}, 15$. In Figure 5(g), action $a_t = 4$ selects the fourth neighbor node 23 of node 15 to add to path $P^l, P^l = \{\{5,17,16,100\}, \{14,8,101\}, \{5,23\}\}$. In Figure 5(h), action $a_t = 2$ selects the second neighbor node 24 of node 23 to add to path $P, P = \{\{5,17,16,100\}, \{14,8,101\}, \{5,23,24\}\}$. Path $P$ now contains the destination node 24, completing the construction of the segmented route.

(3) Reward function design: The reward is the signal that the environment feeds back to the agent after the agent takes action in different states, guiding the agent to build segmented routes. This paper comprehensively considers the multidimensional information in the network state to design reward functions and optimize the segmented routes. The rewards designed in this paper are divided into instantaneous rewards and terminal rewards.

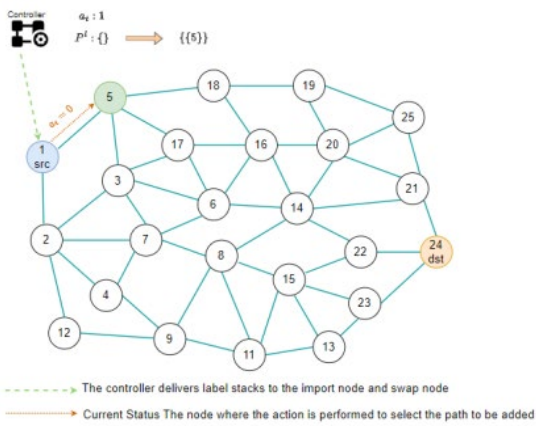

Figure 5 (a)

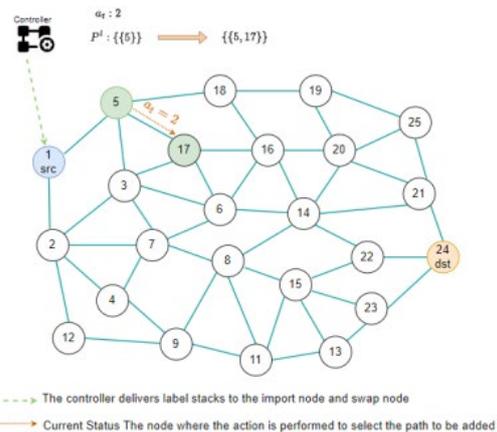

Figure 5 (b)

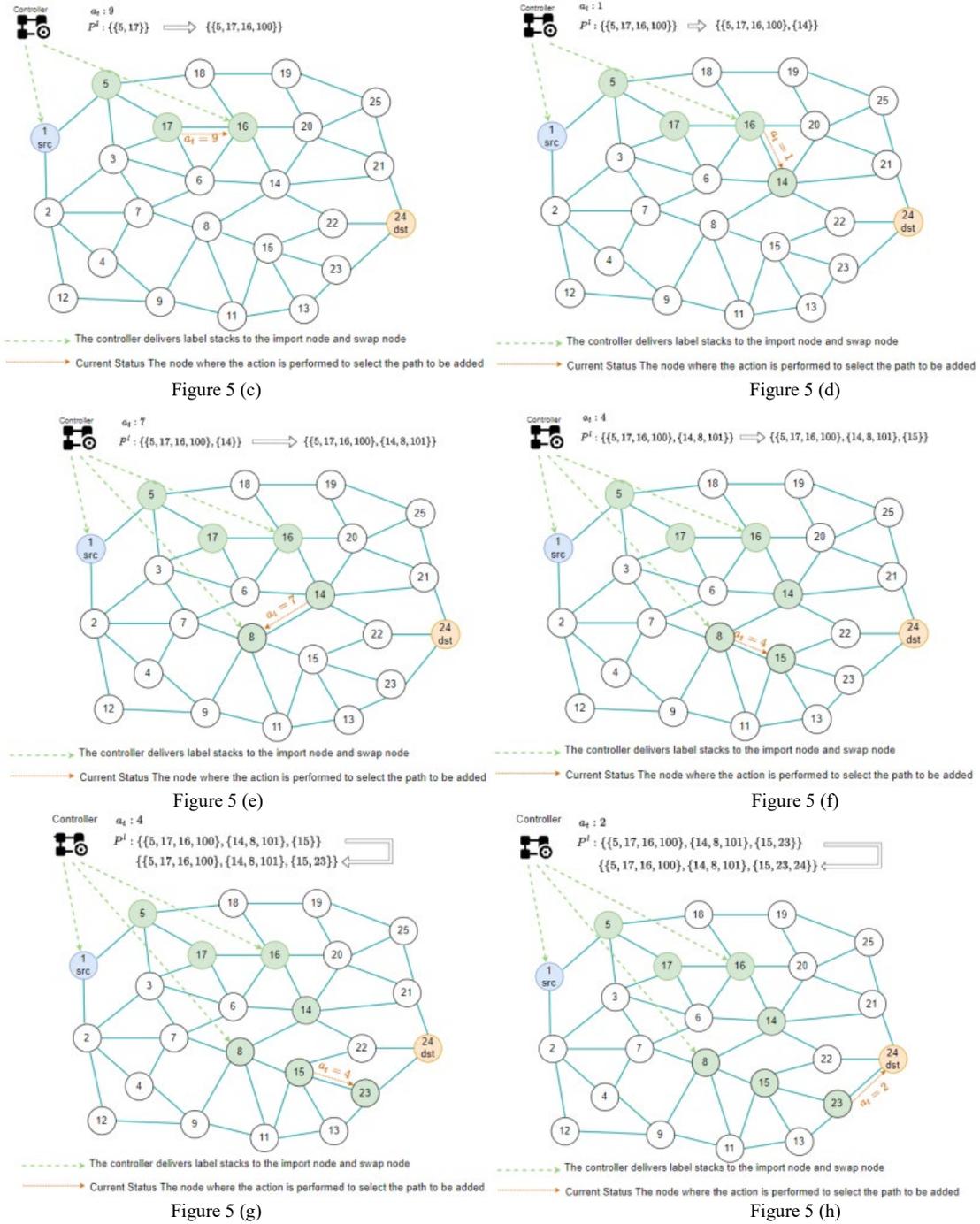

Figure 5 (c)

Figure 5 (d)

Figure 5 (e)

Figure 5 (f)

Figure 5 (g)

Figure 5 (h)

**Figure 5**. Construction of a segmented route.

An instantaneous reward is the reward that the agent receives immediately after performing an action in a specific state. When the agent selects the next action as an effective action and the algorithm is not finished, the instantaneous reward is calculated as shown in Equation (32):

$$R_{immediate} = \beta_1(bw_{ij} - 1) + \beta_2(1 - delay_{ij}) + \beta_3(1 - loss_{ij}) + \sigma\beta_4(1 - cdelay_j) \quad (32)$$

$bw_{ij}$, $delay_{ij}$, and $loss_{ij}$ represent the remaining bandwidth, delay and packet loss rates, respectively, between the last node $i$ in the path of the current state $s$ and node $j$, where the action is added to the path. These three rewards guide the agent to find the optimal path in the dynamically changing network state. $cdelay_j$ indicates the delay from node $j$ executing the action to the switch. The four parameters are normalized to the range of [0, 1] via the Max-Min method. $\sigma \in {0,1}$, $\sigma = 1$ if the node selected to join the path is a swap node; otherwise, $\sigma = 0$. This parameter guides the agent to determine the optimal swap node in the path. $\beta_k \in [0, 1], k = 1,2,3,4$ are the weight factors of the remaining bandwidth $bw_{ij}$, delay $delay_{ij}$, packet loss $loss_{ij}$ and delay $cdelay_j$ of the network link between nodes $i$ and $j$. When the agent selects the next action as an invalid action, the instantaneous reward is $R_{immediate} = -2$. A large, fixed penalty value is given here to prevent the agent from choosing these invalid actions. When the agent reaches the destination node, the final reward is $R_{end} = 10$. A large positive reward is used to guide the agent to find the destination node, at which time the algorithm terminates and the agent completes the construction of the segmented route.

(4) Updating the SAC network parameters: In the SAC algorithm, two action value functions $Q$ (parameters $w_1$ and $w_2$) and a policy function $\pi$ (parameter $\theta$) are modeled. SAC uses two $Q$ networks, but each time it uses a $Q$ network, it selects a $Q$ network with a small value, thereby alleviating the problem of overestimating $Q$ values. The loss function of any $Q$ function is shown in Equation (33):

$$L_Q(\omega) = E_{(s_t,a_t,r_t,s_t)\sim R}[\frac{1}{2}Q_\omega(s_t,a_t) - (r_t + \gamma V_\omega - s_{t+1}))^2]$$
$$= E_{(s_t,a_t,r_t,s_{t+1})\sim s_t, a_{t+1}\sim \pi_\theta(?|s_{t+1})}[\frac{1}{2}(Q_\omega(s_t,a_t) - (r_t + \gamma(\min_{j=1,2}Q_{\omega_j} - (s_{t+1},a_{t+1}) - \alpha\log\pi(a_{t+1}|s_{t+1}))))^2] \quad (33)$$

where $R$ represents the data that the policy has collected in the past because the SAC is an offline policy algorithm. To stabilize the training process, the $Q$ target network $Q_{w-}$ is used here, which is also a two-target $Q$ network, corresponding to two $Q$ networks. The loss function of strategy $\pi$ is obtained from the KL divergence, which is simplified to Equation (34):

$$L_\pi(\theta) = E_{s_t\sim R, a_t\sim \pi_\theta}[\alpha\log(\pi_\theta(s_t|a_t))Q_\omega(s_t,a_t)] \quad (34)$$

This method can be viewed as maximizing the function $V$:

$$V(s_t) = E_{a_t\sim \pi_\theta}[Q_\omega(s_t,a_t) - \alpha\log(\pi(a_t,s_t))] \quad (35)$$

In the environment of a continuous action space, the strategy of the SAC algorithm outputs the mean and standard deviation of the Gaussian distribution, but the process of sampling the action according to the Gaussian distribution is not derivable. Therefore, reparameterization is needed. The reparameterization method first samples from a unit Gaussian distribution $\mathcal{N}$ and then multiplies the sample value by the standard deviation and adds the mean value. This method can be viewed as sampling from a policy Gaussian distribution and is derivable for the policy function. It is expressed as $a_t = f_\theta(\in_t; s_t)$, where $\in_t$ is a noisy random variable. The loss function of the rewriting strategy accounts for two functions $Q$ simultaneously:

$$L_\pi(\theta) = E_{s_t\sim R, \in_t\sim N}[\alpha\log(\pi_\theta(f_\theta(\in_t;s_t)|s_t) - \min_{j=1,2}Q_{\omega_j}(s_t, f_\theta(\in_t;s_t))] \quad (36)$$

In the SAC algorithm, choosing the coefficient of the entropy regular term is highly important. Different entropies are required in different states: in a state where the optimal action is uncertain, the entropy value should be larger; however, in a state where the optimal action is relatively certain, the entropy value can be smaller. To automatically adjust the entropy regular term, SAC rewrites the objective of reinforcement learning into a constrained optimization problem, as shown in Equation (37):

$$\max_\pi \mathbb{E}_\pi[\sum_t r(s_t,a_t)] s.t. \mathbb{E}_{s_t,a_t\sim \rho_\pi}[-\log(\pi_t(a_t|s_t))] \geq H_0 \quad (37)$$

To maximize the expected return, the mean constraint entropy is greater than $\mathcal{H}_0$. After simplification via several mathematical techniques, the loss function of $\alpha$ is obtained:

$$L(\alpha) = E_{s_t\sim R, a_t\sim \pi(\cdot|s_t)}[-\alpha\log\pi(a_t|s_t) - \alpha H_0] \quad (38)$$

When the entropy of the policy is lower than the target value $\mathcal{H}_0$, the $\alpha$ value of the training target $L(\alpha)$ increases, and the importance of the corresponding term of the policy entropy increases in the above minimization of the loss function $L_\pi(\alpha)$. However, when the entropy of the strategy is greater than the target value $\mathcal{H}_0$, the $\alpha$ value of the training target decreases, which focuses the strategy training on value enhancement.

*4.3. DRL-SR algorithm flowchart*

The implementation of the DRL-SR algorithm framework is shown in Algorithm 1, which finds the optimal segmentation routing path $P$ from source node $src$ and destination node $dst$ from the current network topology $G(V,E,W)$. The input of the algorithm includes the environment $s$ of the SAC algorithm, the learning rate $\alpha$ of the agent, the network parameter update frequency $f_{update}$, the size $k$ of the batch collected each time from the experience pool $R$, and the algebra $M$ of the training. The optimal segmented route $P$ from the source node to the destination node is the output. Lines 1 through 3 initialize the entire DRL network and the experience playback pool. The fourth line is the cycle of the number of training rounds. Line 5 reads the topology $G(V,E,W)$ from the NLI repository. Line 6 initializes the DRL state $s$. In lines 7 to 10, agents explore according to strategy $\pi$, interact with the environment and collect $(s_t, a_t, r_t, s_{t+1})$ into the experience pool. Lines 11 to 17 update parameters according to the SAC algorithm. Finally, the agent learns to build the segmented routing $P$ strategy $\pi_\theta(s)$.

---

Algorithm 1: **DRL-SR**

**Input:** network topology $G(V, E, W)$, SAC algorithm environment $s$, learning rate of the agent $\alpha$、neural network parameter update frequency $f_{update}$、size of each batch collected from the experience pool $N$、trained $Episode$.
**Output: o**ptimal segmented route path from the source node to the destination node $P$.

1: parameter $w_1$, $w_2$ and $\theta$ initialize Critic networks separately $Q_{\omega 1}(s, a)$ and $Q_{\omega 2}(s, a)$, Actor networks $\pi_\theta(s)$
2: copy the same parameters $w_1^- \leftarrow w_1, w_2^- \leftarrow w_2$, and initialize the target network separately $Q_{w1^-}$ and $Q_{w2^-}$
3: initialize the experience playback pool $R$;
4: **for** $e = 1 \rightarrow Episode$ **do**
5:    **for** $TM$ in NLI stash **do**
6:       $(src, dst)$ initial state of environment $s_1$
7:       **for** $t = 1 \rightarrow T$ **do**
8:          select an action based on the current policy $a_t = \pi_\theta(s_t)$
9:          execution action $a_t$, get bonus value $r_t$, environmental status becomes $s_{t+1}$
10:         $(s_t, a_t, r_t, s_{t+1})$ storage playback pool $R$
11:         **for** train episode $k = 1 \rightarrow K$ **do**
12:            $R$ Sample $N$ tuples $\{(s_t, a_t, r_t, s_{t+1})\}_{i=1,\cdots,N}$
13:            for each tuple, the target network is used $y_i = r_i + \gamma \min_{j=1,2} Q_{\omega_j^-}(s_{t+1}, a_{t+1}) - \alpha \log(\pi_\theta(a_{i+1} \mid s_{i+1}))$
14:            make the following updates to both Critic networks: $j = 1,2$, minimization loss function $L = \frac{1}{N}\sum_{i=1}^{N}\left(y_i - Q_{\omega_j}(s_i, a_i)\right)^2$
15:            sample action $\tilde{a}_i$ with the reparameterization technique, update the current Actor network with the following loss function $L_\pi(\theta) = \frac{1}{N}\sum_{i=1}^{N}\left(\alpha \log(\pi_\theta(\tilde{a}_i \mid s_i)) - \min_{j=1,2} Q_{\omega_j}(s_i, \tilde{a}_i)\right)$
16:            Updating the coefficient $\alpha$ of the entropy regular term
17:            update target network: $w_1^- \leftarrow \tau w_1 + (1 - \tau)w_1^-, w_2^- \leftarrow \tau w_2 + (1 - \tau)w_2^-$
18:         **end for**
19:      **end for**
20:   **end for**
21: **end for**
22: The agent learns to construct the segmented routing $P$ policy.

## 5. Experimental setup and performance evaluation

### 5.1. Simulation environment setup

In this paper, the simulation network topology is built via the Mininet 2.3.0 simulation environment platform. Mininet allows users to create complex and highly customized SDN environments. The entire simulation environment is deployed on an Ubuntu 20.04.6 server with a GeForce RTX 3090 graphics card. The user datagram protocol (UDP) packets are sent between nodes in the network via Iperf to simulate real network traffic.

Ryu is used as the SDN controller and is responsible for the response of events such as node flow table delivery. Ryu collects network information and stores it in Pickle files in graph format to generate network traffic datasets. Finally, Python 3.6 and PyTorch 1.11.0 are used to realize the interaction between the SDN and DRL.

To simulate rich and varied network resources, three real network topologies are introduced in this paper as test scenarios. The three topologies are from the Internet Topology Zoo (http://www.topology-zoo.org/), namely, ARPANet, Sun, and Arnes. The network topology link parameters are randomly generated and uniformly distributed. The three topologies are shown in Figure 6(a), Figure 6(b), and Figure 6(c), respectively. The random ranges of the link bandwidth and delay in the network topology are 500 Mbps and 1 to 10 ms, respectively.

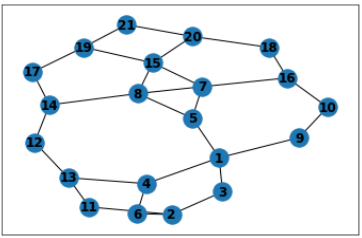
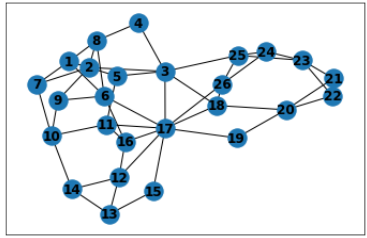
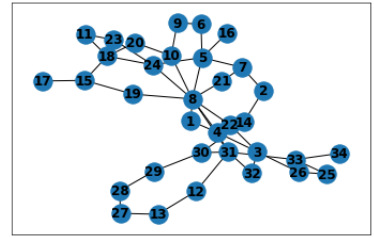

**Figure 6**(a). ARPANet topology      **Figure 6**(b). Sun topology      **Figure 6**(b). Arnes topology

This paper simulates the network traffic situation 24 hours a day, as shown in Figure 7. The horizontal coordinate is the time, and the vertical coordinate is the average traffic sent by each node in Mbit/s, which corresponds to the distribution of network traffic at different times of the day.

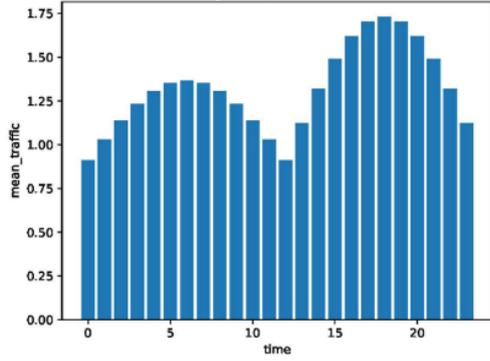

**Figure 7**. Simulated flow distribution map

*5.2. Performance index*

The design objective of this paper is to optimize the routing path and the label stack division process in the segmented routing scheme. For the path performance evaluation, the minimum remaining bandwidth, delay and packet loss rate of the path are used to evaluate the path performance, and the average values of the minimum remaining bandwidth, delay and packet loss rate at different times are used to represent the network performance according to the simulated network traffic situation. For label stack partitioning, according to the simulated network traffic, the path establishment delays at different times are used as the evaluation index. The smaller the path establishment delay is, the shorter the establishment time of the segmented routes is.

*5.3. Deep reinforcement learning parameter setting*

The setting of hyperparameters affects the performance and convergence speed of the agent. The influence of different hyperparameters on the agent is analyzed, and the optimal hyperparameters are selected.

Batch_size is the number of samples taken each time the model is trained. Batch_size can affect the convergence speed of the model and the final performance of the model. Generally, when the batch size is small, fewer samples are used in each iteration, and each training iteration contains more sample information, which can help the model move out of the local optimum and improve the generalization performance of the model but may lead to increased noise in gradient estimation. A larger Batch_size usually results in faster convergence of the model, and reducing the noise of the gradient estimation helps in more stable convergence, so the learning rate needs to be adjusted appropriately to stabilize the training. Different Batch_size results are shown in Figure 8.

The experimental results show that when Batch_size is set to 16, the convergence rate of the rewards obtained by the agent is the slowest. When set to 128, the initial reward value converges to a local optimum, and the training algebra reaches its maximum value at approximately 700. A Batch_size of 64 works best for agents.

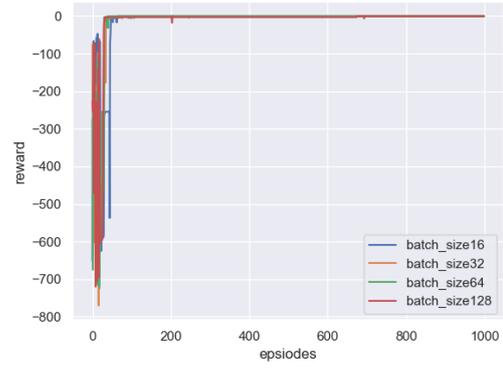

**Figure 8**. Learning curve on different batchsize

The learning rate is an important hyperparameter that controls the updating amplitude of the model parameters in each iteration. Too high a learning rate may cause the model to fail to converge, whereas too low a learning rate may cause the model to converge very slowly or fail to learn. In this paper, the SAC algorithm framework is used for DRL. There are two neural networks: Actor and Critic. The learning rate of one network is fixed, and the learning rate of the other network is adjusted. First, the Critic network learning rate is fixed at $\alpha_2 = 1e-3$, the Actor network learning rate is adjusted to $\alpha_1$, and the results are shown in Figure 9.

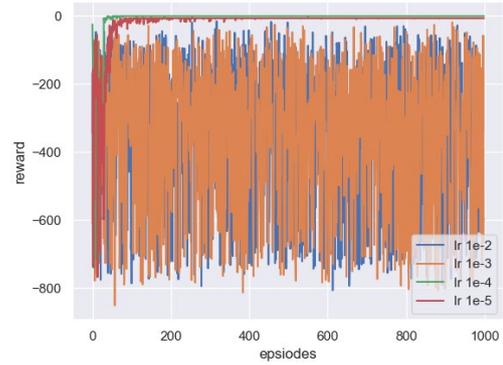

**Figure 9**. Learning curve of learning rate $\alpha_1$

The experimental results show that when $\alpha_1 = 1e-2$ and $\alpha_1 = 1e-3$, the reward value obtained by the agent with a higher learning rate has difficulty converging. When $\alpha_1 = 1e-4$ and $\alpha_1 = 1e-5$, the reward value can converge, but when $\alpha_1 = 1e-4$, the convergence rate is faster than when $\alpha_1 = 1e-5$, and the reward value is larger. The fixed Actor network learning rate is $\alpha_1 = 1e-4$, and the adjusted Critic network learning rate is $\alpha_2$. The results are shown in Figure 10.

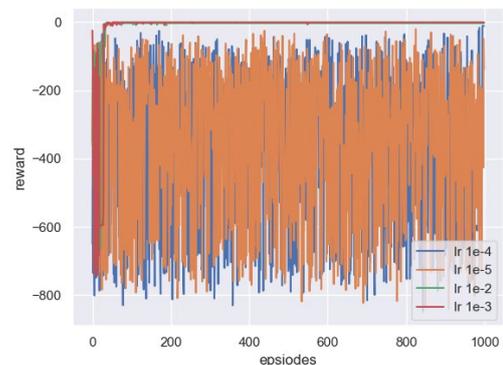

**Figure 10**. Learning curve of learning rate $\alpha_2$

Setting the learning rate of the critic network at $\alpha_2 = 1e-4$ and $\alpha_2 = 1e-5$, the reward value obtained by the agent has difficulty converging. When $\alpha_2 = 1e-2$ and $\alpha_2 = 1e-3$, the reward value can converge, but when $\alpha_2 = 1e-3$, the reward value is larger.

*5.4. Contrast experiment*

To evaluate the performance of the DRL-SR algorithm, we set the fixed source node to the destination node in the 10-node, 14-node and 21-node wireless network topologies to simulate network traffic in the real world and compare and analyze the network performance of the DRL-SR algorithm and traditional OSPF. The performance indicators evaluated are the minimum remaining bandwidth, delay, and packet loss rate of the path. Moreover, the path establishment speeds of the DRL-SR and common SR segment routes are compared and analyzed.

Figure 11(a), 11(b), and 11(c) show the average bandwidths of the total link bottleneck of the path from the source node to the destination node of the indicator agent. The average path throughput obtained by the DRL-SR algorithm in this paper is improved by 5.15%, 6.12% and 9.79% on average compared with those of the OSPF algorithm in the three topologies. The path selected by the algorithm in this paper has a larger bottleneck bandwidth and can transmit more data, meeting the performance requirements of data transmission.

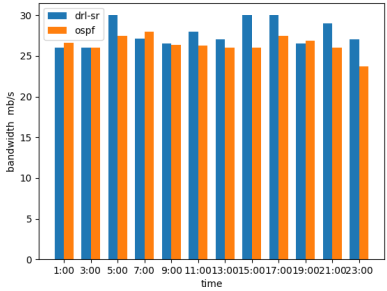
**Figure 11(a)**. ARNet bottleneck bandwidth results

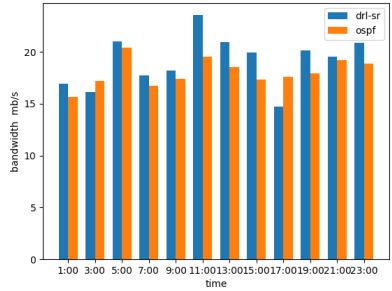
**Figure 11(b)**. Sun bottleneck bandwidth results

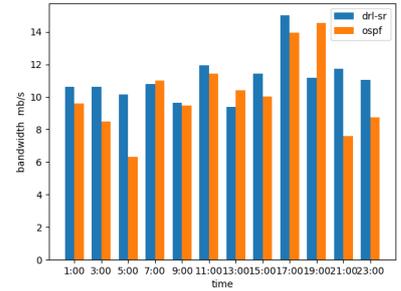
**Figure 11(c)**. Arnes. bottleneck bandwidth results

Figure 12(a), 12(b), and 12(c) show the average values of the total link delays of the path from the source node to the destination node. The average path delays obtained by the DRL-SR algorithm in this paper are reduced by 21.04%, 23.43% and 6.39% on average compared with those of the OSPF algorithm for the three topologies. Experiments show that the proposed algorithm is more inclined to find the path with the lowest network delay and satisfied the performance requirements concerning low network delays.

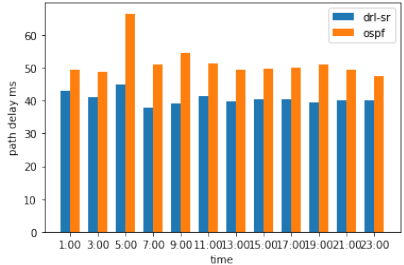
**Figure 12(a)**. ARNet delay results

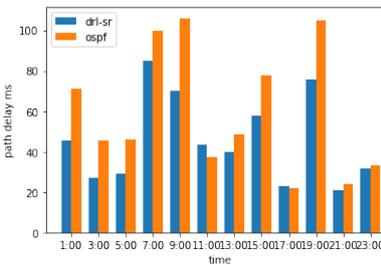
**Figure 12(b)**. Sun delay result

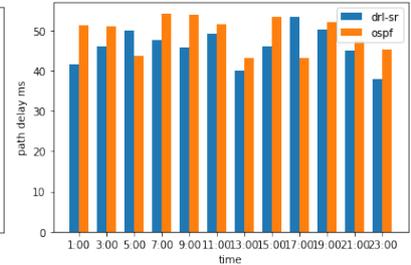
**Figure 12(c)**. Arnes topology delay result

The measurement indicators in Figure 13(a), 13(b), and 13(c) are the average packet loss rates of the path from the source node to the destination node. The packet loss rate of the DRL-SR algorithm is significantly lower than that of the OSPF algorithm. Packet loss occurs at a certain probability for each link. In particular, when the network traffic increases, the packet loss rate increases. The experimental results show that the bottleneck bandwidth of the selected path is larger than that of other algorithms and can meet the transmission demand of large amounts of traffic, so it can effectively avoid packet loss.

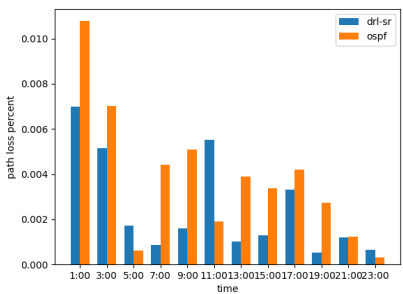
**Figure 13(a)**. ARNet packet loss rate results

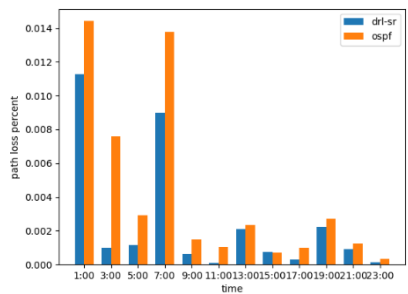
**Figure 13(b)**. Sun packet loss rate result

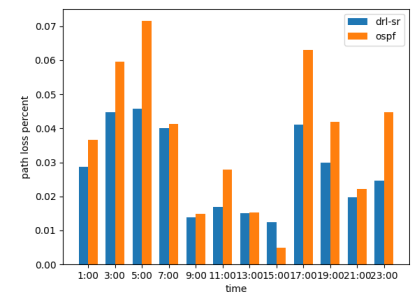
**Figure 13(c)**. Arnes packet loss rate result

The measurement indicators in Figure 14(a), 14(b), and 14(c) are the latency values induced when delivery the flow table. Compared with those of the common segmented routing label stack division scheme, the runoff table-based delivery delays obtained by the DRL-SR algorithm in this competition are reduced by 9.6%, 19.93% and 7.13% on average for the three topologies. Experiments show that the label stack partition of this algorithm can accelerate the flow table delivery speed and better dynamically adjust the routing strategy.

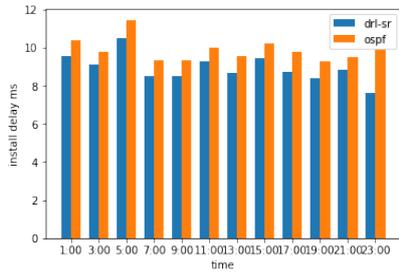
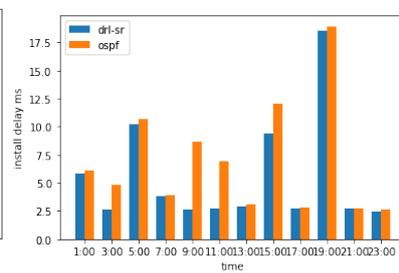
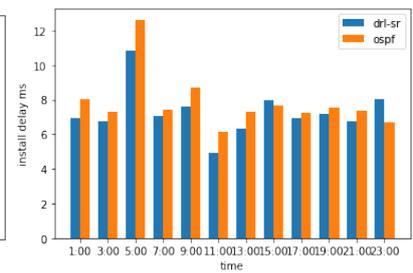

**Figure 14(a).** ARNet delay results of flow table delivery  **Figure 14(b).** Sun delay results of flow table delivery  **Figure 14(c).** Arnes delay results of flow table delivery

## 6. Conclusion

In this paper, an intelligent segmented routing method based on deep reinforcement learning (DRL-SR) is proposed. To adjust routing policies in the dynamic network of traditional SDNs, it is necessary to reissue the flow table for all paths, which leads to slow network convergence, and many traditional algorithms rely on local information to route decision making, which easily leads to suboptimal global results. The DRL-SR algorithm can constantly learn and adjust in a dynamic network environment. In accordance with the update strategy of the network state, routing and forwarding paths with larger bandwidths, shorter delays and lower packet loss rates are constructed. More importantly, during the process of establishing a routing and forwarding path, it is not necessary to update the flow table for all nodes in the path, only for some nodes in the path with low communication delay between the controller and the path. This approach ensures the fastest flow table delivery, accelerates network convergence, reduces data loss caused by path switching, and improves the overall performance of the network.

**Acknowledgements**

This work was supported in part by the National Natural Science Foundation of China (Nos.62161006, 62372353), Key Laboratory of Cognitive Radio and Information Processing of Ministry of Education (No. CRKL220103), and the subsidization of the Innovation Project of Guangxi Graduate Education (No. YCSW2023134).

**Conflicts of Interest**

The authors declare that they have no known competing financial interests or personal relationships that could have appeared to influence the work reported in this paper. The authors also declare no conflicts of interest.

**Data Availability Statement**

The authors confirm that the data supporting the findings of this study are available within the article. Data sharing may be applicable to this article when requested.